\documentclass[review]{elsarticle}
\usepackage[a4paper, scale=0.8]{geometry}

\usepackage{lineno,hyperref}

\usepackage{float}

\usepackage{colortbl}
\usepackage{xcolor}
\usepackage{adjustbox}
\usepackage{booktabs}
\usepackage{multirow}
\usepackage{bm}
\usepackage{amssymb}
\usepackage{dsfont}
\usepackage{amsmath}
\usepackage{pifont}
\usepackage{subfig}
\usepackage{algorithm}
\usepackage{algorithmic}

\journal{arXiv}

\begin{document}

\begin{frontmatter}



\title{Dynamic Prompt Allocation and Tuning for Continual Test-Time Adaptation}


\author[1]{Chaoran~Cui}
\ead{crcui@sdufe.edu.cn}
\author[1]{Yongrui~Zhen}
\ead{yrzhen2000@163.com}
\author[1]{Shuai~Gong}
\ead{gsh8210@163.com}
\author[1]{Chunyun~Zhang}
\ead{zhangchunyun1009@126.com}
\author[1]{Hui~Liu}
\ead{liuh_lh@sdufe.edu.cn}
\author[2]{Yilong~Yin}
\ead{ylyin@sdu.edu.cn}

\address[1]{School of Computing and Artificial Intelligence, Shandong University of Finance and Economics, Jinan, China}
\address[2]{School of Software, Shandong University, Jinan, China}

% \cortext[cor1]{Corresponding author: Chaoran Cui}

\begin{abstract}
Continual test-time adaptation (CTTA) has recently emerged to adapt a pre-trained source model to continuously evolving target distributions, which accommodates the dynamic nature of real-world environments.
To mitigate the risk of catastrophic forgetting in CTTA, existing methods typically incorporate explicit regularization terms to constrain the variation of model parameters.
However, they cannot fundamentally resolve catastrophic forgetting because they rely on a single shared model to adapt across all target domains, which inevitably leads to severe inter-domain interference.
In this paper, we introduce learnable domain-specific prompts that guide the model to adapt to corresponding target domains, thereby partially disentangling the parameter space of different domains.
In the absence of domain identity for target samples, we propose a novel dynamic Prompt AllocatIon aNd Tuning (PAINT) method, which utilizes a query mechanism to dynamically determine whether the current samples come from a known domain or an unexplored one.
For known domains, the corresponding domain-specific prompt is directly selected, while for previously unseen domains, a new prompt is allocated.
Prompt tuning is subsequently performed using mutual information maximization along with structural regularization.
Extensive experiments on three benchmark datasets demonstrate the effectiveness of our PAINT method for CTTA. 
We have released our code at \url{https://github.com/Cadezzyr/PAINT}.
\end{abstract}



\end{frontmatter}

% \linenumbers

\section{Introduction}
Deep neural networks have demonstrated considerable success across a broad spectrum of visual recognition tasks.
However, their impressive effectiveness is generally observed when the training data and test data share the same distribution.
In real-world applications, we often encounter distribution shifts, posing a substantial barrier to the generalization of deep models.
Consequently, numerous research efforts have been devoted to enhancing model robustness against distribution shifts.
For example, domain adaptation (DA)~\cite{wilson2020survey} aims to transfer knowledge learned from a labeled source domain to an unlabeled target domain, while domain generalization (DG)~\cite{zhou2023domain} focuses on leveraging data from multiple source domains to learn a model that can generalize well to any out-of-distribution target domain.

Both DA and DG methods require access to labeled source data.
This requirement may be impractical due to privacy concerns and limitations in storage resources.
We may only have access to a pre-trained source model, with the expectation that this model can effectively adapt to unlabeled target data at test time.
This scenario is commonly referred to as test-time adaptation (TTA)~\cite{liang2023comprehensive}.
TTA often needs to be conducted in an online manner, where target data is processed sequentially, and the model adapts to each new batch of data before making predictions.
In online TTA, the model is prohibited from storing target data during adaptation~\cite{su2024revisting}.

Traditional TTA research~\cite{wang2021tent,iwasawa2021test,chen2022contrastive} assumes that all target samples originate from the same stationary domain.
In practice, however, the distribution of target samples can continuously change over time.
For instance, an autonomous driving system must navigate through diverse road conditions, cope with varying weather, and handle unexpected scenarios.
To accommodate this situation, continual test-time adaptation (CTTA)~\cite{wang2022continual,niu2022efficient} has been introduced, where the target test data is streamed from an environment that continuously evolves.

% A major challenge in CTTA is known as \emph{catastrophic forgetting}~\cite{kirkpatrick2017overcoming}: when the model sequentially performs adaptation tasks from dynamic data distributions, it may forget the knowledge it had previously acquired.
% The major challenges in CTTA arise from two aspects: 1) catastrophic forgetting~\cite{kirkpatrick2017overcoming} and 2) error accumulation~\cite{song2023ecotta}.
% Catastrophic forgetting occurs when a model forgets the knowledge it had previously acquired while sequentially performing adaptation tasks from dynamic data distributions.
% To mitigate catastrophic forgetting, previous studies~\cite{wang2022continual,niu2022efficient} typically include an additional regularization term to prevent the model's parameters from drastic changes.
A major difficulty in CTTA is catastrophic forgetting~\cite{kirkpatrick2017overcoming}, where the model forgets previously acquired knowledge as it continuously adapts to dynamic data distributions.
To mitigate catastrophic forgetting, prior research typically incorporates an additional regularization term to constrain the variation of model parameters.
For example, CoTTA~\cite{wang2022continual} proposes to stochastically restore a small portion of parameters to their initial values from the pre-trained source model to help preserve source knowledge over the long term.
EcoTTA~\cite{song2023ecotta} utilizes a self-distillation strategy to ensure that the output of the adaptation model does not deviate significantly from that of the source model.
EATA~\cite{niu2022efficient} introduces a Fisher regularizer to prevent important model parameters from undergoing drastic changes.
% Error accumulation refers to the compounding of small errors introduced during each adaptation step, which over time leads to significant performance degradation in CTTA.
% Prior research~\cite{wang2021tent,shu2022test} usually applies the entropy minimization principle to alleviate error accumulation, encouraging the model to make more confident predictions.

% Despite impressive progress, we argue that existing methods may fail to sufficiently address the issues of catastrophic forgetting and error accumulation because they rely on a single shared model to learn all adaptation tasks, which is subject to severe \emph{inter-task interference}~\cite{wang2024comprehensive}.
Despite remarkable progress, existing methods may still fail to sufficiently address the issue of catastrophic forgetting.
Their limitation lies in relying on a single shared model to adapt across all target domains, which inevitably introduces severe \emph{inter-domain interference}~\cite{wang2024comprehensive}.
When the shared model adapts to new target domains, its parameters are updated to optimize the performance on the new data.
This process risks overwriting parameters that were crucial for previous domains, leading to catastrophic forgetting.
Therefore, to fundamentally resolve catastrophic forgetting, it is essential to allow different domains to be learned in \emph{a separated way} for CTTA.

%\emph{catastrophic forgetting}~\cite{kirkpatrick2017overcoming}: when the model sequentially performs adaptation tasks from dynamic data distributions, it may forget the knowledge it had previously acquired.
%To mitigate catastrophic forgetting, earlier studies on CTTA typically include an additional regularization term to constrain the model's parameters.
%For example, CoTTA~\cite{wang2022continual} suggests stochastically restoring a small portion of parameters to their initial values from the pre-trained source model to help preserve source knowledge in the long-term.
%EATA~\cite{niu2022efficient} introduces a Fisher regularizer to prevent important model parameters from drastic changes.

%In the filed of continual learning~\cite{wang2024comprehensive}, a common solution to catastrophic forgetting is experience replay, which retains some historical data within a memory buffer and optimizes the model using a mix of current data and historical data during each iteration.
%However, experience replay is not suitable for CTTA, because the model is generally prohibited from storing the data while adapting to each batch of target samples~\cite{Doebler2023robust}.

In this paper, we draw inspiration from recent advances in prompt tuning for deep neural networks~\cite{Jia2022,Wang2022}.
Prompt tuning involves inserting a few learnable prompt tokens as extra inputs to facilitate the rapid adaptation of a pre-trained model to downstream tasks.
By designing domain-specific prompts for individual target domains, we can partially disentangle the parameter space across different domains.
Samples from a target domain contribute solely to optimizing their corresponding domain-specific prompts.
In this way, the interference between domains can be significantly reduced, thereby minimizing catastrophic forgetting.
However, realizing this idea in the context of CTTA presents a significant challenge: \emph{the domain identity of target samples is unknown.}
This uncertainty makes it difficult to determine which adaptation task is currently being processed and hinders the automatic selection of an appropriate prompt.

To tackle this challenge, we propose a novel dynamic Prompt AllocatIon aNd Tuning (PAINT) method for CTTA.
Specifically, we maintain a memory buffer to store all domain-specific prompts, with each prompt organized as a key-value pair.
Initially, the memory buffer is empty and gradually expands as the target samples arrive.
For each batch of target samples, we employ a query mechanism to dynamically determine whether the data comes from a known domain or an unexplored one.
For known domains, we directly select the corresponding domain-specific prompt from the memory buffer.
In cases of unseen domains, we allocate a new prompt in the memory buffer for that domain.
Then, we perform prompt tuning following the principle of mutual information maximization~\cite{liang2022source}, which improves both individual certainty and global diversity of the target predictions.
Structural regularization is further incorporated to ensure consistency between the predictions on interpolated target samples and the interpolated sample labels~\cite{zhang2018mixup}, thereby enhancing the model's generalization ability in the target domain.
% enhance the model's generalization ability in the target domain, which enforces consistency between the predictions on interpolated target samples and the interpolated sample labels~\cite{zhang2018mixup}.
Beyond domain-specific prompts, we also fine-tune the shallow encoder blocks of the pre-trained model, which can be regarded as the common knowledge shared by all domains.
Finally, the prompt key is updated using a simple moving average strategy.

It is important to note that, although some studies~\cite{Wang2022,gan2023decorate} have explored prompt tuning in continual learning, they either predetermine the number of prompts or require explicit boundaries between different domains, which significantly differ from our PAINT method.
To sum up, our main contributions are as follows:
\begin{itemize}
  \item We recognize that catastrophic forgetting essentially stems from the inter-domain interference in CTTA. 
In contrast to prior research that rely on a single shared model to adapt across all target domains, we introduce domain-specific prompts to guide model adaptation, enabling the parameter space of different domains to be partially disentangled.
  \item Given the absence of domain identity for target samples, we design a query mechanism that dynamically determines whether to select an existing domain-specific prompt or to allocate a new prompt for the current samples.
Prompt tuning is carried out using mutual information maximization along with structural regularization.
  \item Extensive experiments demonstrate the effectiveness of our PAINT method for CTTA, which achieves the state-of-the-art performance on three benchmark datasets.
\end{itemize}

The remainder of the paper is structured as follows:
Section~\ref{sec:related_work} provides a review of related work.
Section~\ref{sec:preliminary} presents the preliminary knowledge of our study.
Section~\ref{sec:method} details our PAINT method for CTTA.
Experimental results are discussed in Section~\ref{sec:experiments}, followed by the conclusions in Section~\ref{sec:conclusions}.

\section{Related Work}\label{sec:related_work}
In this section, we first review the existing literature on TTA.
Then, we present a brief overview of two closely related fields: continual learning and prompt tuning.

\subsection{Test-Time Adaptation}
TTA aims to adapt a pre-trained model from the source domain to unlabeled data in the target domain~\cite{liang2023comprehensive}.
Fully TTA, which is often known as source-free domain adaptation~\cite{liang2022source}, leverages all the target data to perform adaptation before making any predictions.
In contrast, online TTA~\cite{su2024revisting} incrementally adapts the model as each target sample or small batch of samples is encountered.
Numerous TTA methods have explored techniques such as self-training and entropy minimization for adapting the source model.
For example, TENT~\cite{wang2021tent} updates the model parameters in batch normalization layers via entropy minimization.
CPL~\cite{goyal2022test} proposes a conjugate pseudo labeling strategy to enhance self-training on the target domain.

CTTA refers to a scenario where the target domain undergoes continuous changes, presenting additional challenges for conventional TTA methods, particularly the risk of catastrophic forgetting.
As the pioneering work, CoTTA~\cite{wang2022continual} introduces a teacher-student pseudo-labeling framework augmented with stochastic weight restoration.
RMT~\cite{Doebler2023robust} further proposes to use symmetric cross-entropy as the consistency loss between the teacher and student models.
EATA~\cite{niu2022efficient} and SAR~\cite{niu2023towards} select reliable target samples and only use them to update the model for test-time adaptation.
SANTA~\cite{Chakrabarty2023santa} leverages a self-distillation technique that regards the source model as an anchor to align target features.

Unlike these methods that rely on a shared model for all target domains, we exploit learnable domain-specific prompts to guide model adaptation, partially disentangling the parameter space across different domains.

\subsection{Continual Learning}
Continual learning (CL)~\cite{wang2024comprehensive}, also known as lifelong learning, refers to the ability of a machine learning model to learn sequentially from a stream of data over time, without forgetting previously acquired knowledge.
% Unlike traditional static learning settings, CL scenarios involve dynamic environments where data distributions evolve incrementally. 
% A key difficulty in CL is catastrophic forgetting, where the adaptation to new tasks or data leads to the loss of prior knowledge.
To address catastrophic forgetting, existing CL methods can be broadly classified into three categories: 
regularization-based methods~\cite{kirkpatrick2017overcoming,chaudhry2018riemannian}, which constrain updates to model parameters critical for earlier tasks; 
replay-based methods~\cite{chaudhry2019tiny, shin2017continual}, which replay stored or generated data from past tasks to maintain memory; 
and architecture-based methods~\cite{rusu2016progressive,mallya2018piggyback}, which allocate separate parts of the model to individual tasks to prevent interference. 
Our PAINT method falls into architecture-based methods, but it introduces a novel query mechanism for dynamically allocating domain-specific prompts rather than permanently assigning fixed model components to individual tasks.

Recent studies, including L2P~\cite{Wang2022} and DualPrompt~\cite{wang2022dualprompt}, have utilized learnable prompts to guide models in CL.
PAINT distinguishes itself from these methods in two key aspects:
1) While these methods operate in a supervised manner, PAINT addresses the TTA problem by adapting the source model using unlabeled target samples; and 2) In contrast to these methods, which rely on a fixed number of prompts, PAINT flexibly adjusts the number of prompts as needed.

% The main idea of architecture-based method is to split the parameters of the model or the architecture of the network and test them according to different splitting guidelines (between individual tasks or between shared and specific). Piggyback\cite{mallya2018piggyback} learn weight mask to adapt a single network to multiple tasks,Progressive Networks\cite{rusu2016progressive} introduces an identical sub-network for each task and allows knowledge transfer from other sub-networks via connections.

\subsection{Prompt Tuning}
As one of the most parameter-efficient fine-tuning techniques, prompt tuning was initially proposed in the field of natural language processing~\cite{liu2023pre}.
By inserting either templated or learnable prompt tokens into textual inputs, it enables the adaptation of pre-trained models to downstream tasks.
CoOp~\cite{zhou2022learning} transforms the input textual contexts into learnable prompts for adapting vision-language models to few-shot image recognition.
VPT~\cite{Jia2022} introduces a small set of visual prompt tokens into the vision transformer (ViT)~\cite{Dosovitskiy2021image}.
Maple~\cite{khattak2023maple} utilizes both text prompts and visual prompts to improve the alignment between paired textual and visual representations.

To facilitate CTTA, a recent study named VDP~\cite{gan2023decorate} introduces pixel-level prompts to decorate input images processed by convolutional neural networks (CNNs).
It also develops domain-specific prompts, but is only applicable to scenarios with clearly defined domain boundaries, where target samples from different domains arrive in a strictly sequential manner~\cite{wang2024comprehensive}.
Our PAINT method performs token-level prompt tuning based on transformer architecture.
Different from VDP, PAINT eliminates the need for explicit domain boundaries by designing a dynamic query mechanism that allows target samples to determine the domains to which they belong.

\section{Preliminaries}\label{sec:preliminary}
In this section, we first formulate the problem of CTTA.
Then, we briefly describe the technique of visual prompt tuning.

\subsection{Problem Definition}
In CTTA, we are provided with a pre-trained model $f_{\bm{\Theta}}$ with parameters $\bm{\Theta}$ that has been trained on a source domain.
Our goal is to adapt the pre-trained model to test samples from continuously evolving target domains in an online manner.
The data distribution varies not only between the source domain and each target domain but also among different target domains.
During adaptation, access to source data is restricted, and each test sample can be utilized only once.

Typically, test samples arrive sequentially in batches.
Let $\mathcal{B}= \{\bm{x}_1, \bm{x}_2, \ldots, \bm{x}_B\}$ denote a batch of samples from a certain target domain, where $B$ is the batch size.
The model needs to be adapted according to $\mathcal{B}$ by updating the parameters $\bm{\Theta}$ with one gradient step, and then make online predictions for each sample $\bm{x}_i \in \mathcal{B}$.
\emph{Note that we neither know the domain identity of test samples nor the total number of target domains.}
As a common practice in CTTA, we consider the $K$-way image classification task and assume that different domains share the same category space.

\subsection{Visual Prompt Tuning}
CNN-based methods have dominated the field of CTTA for a long time. 
With the recent popularity of ViT~\cite{Dosovitskiy2021image} in numerous visual applications, ViT-based approaches~\cite{Liu2024continual} have started to show promising results in CTTA.
We employ the visual prompt tuning technique~\cite{Jia2022} to adapt a pre-trained ViT model to new target domains with minimal computation cost.
Given an input image $\bm{x}_i$, it is first divided into a sequence of fixed-size patches and projected into patch embeddings $\bm{E}_i \in \mathbb{R}^{L_e \times D_e}$, where $L_e$ denotes the sequence length and $D_e$ denotes the embedding dimension.
Then, a learnable prompt $\bm{P} \in \mathbb{R}^{L_p \times D_e}$ that consists of $L_p$ prompt tokens is prepended to the patch embeddings for guiding model adaptation.
% The combined sequence together with an extra learnable classification token [CLS] is fed into ViT and processed by a stack of transformer blocks.
% The combined sequence $[\bm{P}; \bm{E}_i]$ is fed into the ViT, which finally predicts a class probability distribution $\bm{y}_i \in [0,1]^{K}$ for the input $\bm{x}_i$.
Conceptually, ViT can be decomposed into a feature encoder $\mathcal{F}(\cdot)$ comprising a stack of transformer blocks and a softmax classification head $\mathcal{H}(\cdot)$.
The combined sequence $[\bm{P}; \bm{E}_i]$ is fed into the encoder to extract visual features $\bm{f}_i \in \mathbb{R}^{D_f}$, which are finally transformed into a class probability distribution $\bm{p}_i \in [0,1]^{K}$ for the input $\bm{x}_i$:
\begin{equation}\label{eq:vit}
\begin{array}{l}
% {\bm{f}_i} = \left\{ {{\mathcal{T}_l}} \right\}_{l=1}^L \left( {[\bm{P}; \bm{E}_i]} \right),\\
{\bm{f}_i} = \mathcal{F} \left( {[\bm{P}; \bm{E}_i]} \right),\\
{\bm{p}_i} = \mathcal{H} \left( {{\bm{f}_i}} \right).
\end{array}
\end{equation}

\section{Method}\label{sec:method}
\begin{figure*}[t]
	\centering
	\includegraphics[width=0.8\textwidth]{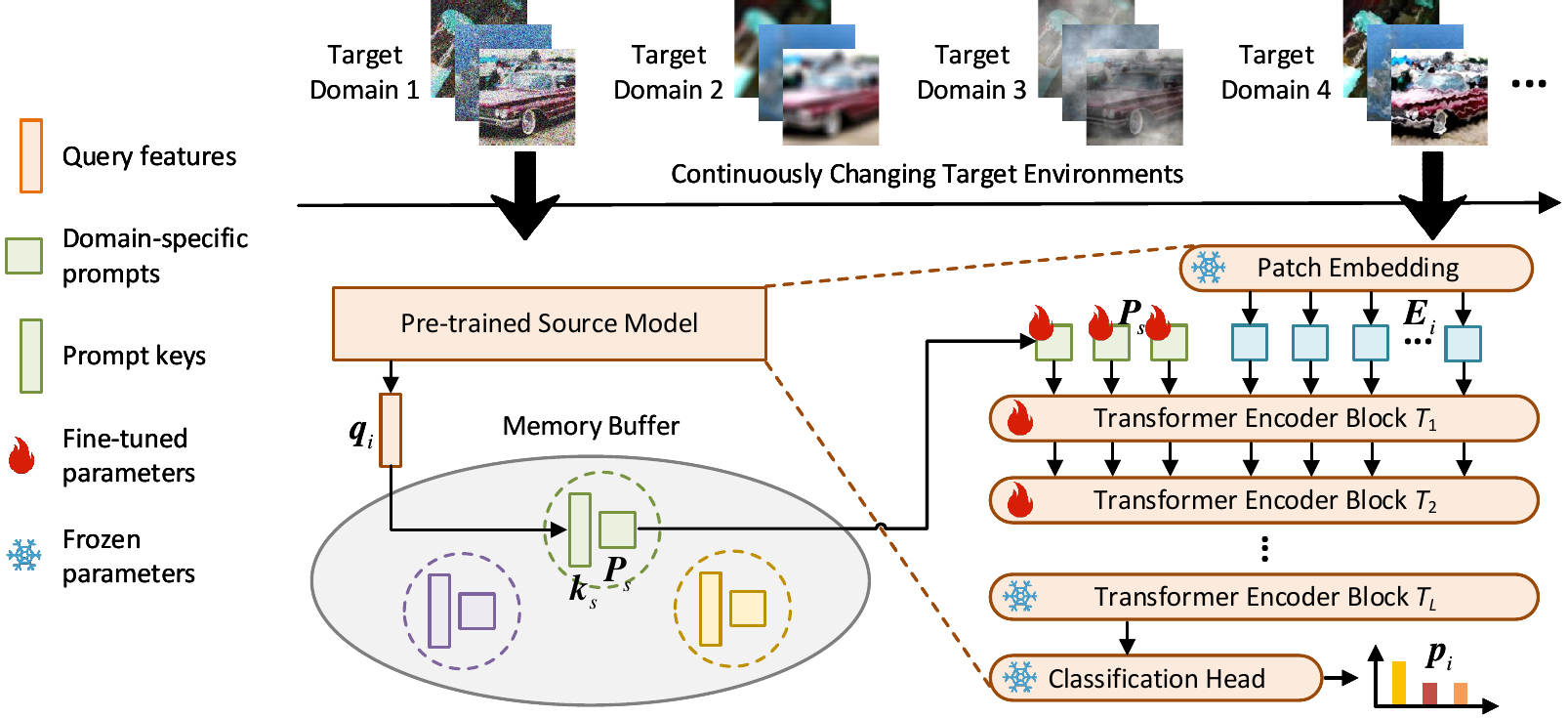}
\caption{Conceptual framework of our PAINT method for CTTA.}
\label{fig:framework}
\end{figure*}

To reduce the interference between domains, we propose a dynamic Prompt AllocatIon aNd Tuning (PAINT) method for CTTA.
PAINT leverages a query mechanism to dynamically allocate a domain-specific prompt for incoming target samples, and integrates mutual information maximization with structural regularization for prompt tuning.
The conceptual framework of PAINT is illustrated in Figure~\ref{fig:framework}.

\subsection{Dynamic Prompt Allocation}\label{sec:prompt_allocation}
The domain identity of target samples is unknown at test time.
To facilitate domain-specific prompt tuning, we aim to empower the samples themselves to decide which prompt to utilize through a query mechanism.
For this purpose, we maintain a memory buffer $\mathcal{M} = \{(\bm{k}_1, \bm{P}_1), (\bm{k}_2, \bm{P}_2), \ldots\}$ to store all domain-specific prompts, in which each prompt $\bm{P}_j$ is regarded as a value and is associated with a key $\bm{k}_j \in \mathbb{R}^{D_f}$.
Initially, the memory buffer $\mathcal{M}$ is empty and gradually expands as the target samples arrive.

Specifically, upon receiving the batch of samples $\mathcal{B}$, we directly adopt the pre-trained source model as a frozen feature extractor to obtain the visual features $\bm{q}_i \in \mathbb{R}^{D_f}$ for each sample $\bm{x}_i \in \mathcal{B}$, i.e., $\bm{q}_i= \mathcal{F} \left( {\bm{E}_i} \right)$.
The sample $\bm{x}_i$ uses $\bm{q}_i$ as a query and retrieves the most appropriate prompt by matching $\bm{q}_i$ with each prompt key $\bm{k}_j$ from the memory buffer. 
In our study, we empirically used the cosine similarity as the matching function, and the retrieval process can be expressed by
\begin{equation}\label{eq:matching}
\bm{k}_s^i, \bm{P}_s^i = {\mathop {\arg \max }\limits_{({\bm{k}_j}, {\bm{P}_j}) \in \mathcal{M}} {\mathop{\rm cosine}\nolimits} ( {\bm{q}_i,{\bm{k}_j}} )},
\end{equation}
where $\mathop{\rm cosine}\nolimits(\cdot,\cdot)$ denotes the cosine similarity function, and $\bm{P}_s^i$ denotes the prompt selected by $\bm{x}_i$ and is paired with the key $\bm{k}_s^i$.
During the prompt allocation, all samples in the batch can make their own choices, and we aggregate their opinions via majority voting to reach a consensus on the optimal prompt for the batch of data:
\begin{equation}\label{eq:voting}
\bm{k}_s, \bm{P}_s = \mathop {\arg \max }\limits_{({\bm{k}_j}, {\bm{P}_j}) \in \mathcal{M}} \sum\limits_{i = 1}^B {\mathds{1}\left( {{\bm{k}_j} = \bm{k}_s^i} \right)},
\end{equation}
where $\mathds{1}(\cdot)$ is an indicator function that outputs 1 if the input argument is true and 0 otherwise.

It is worth noting that the domain-specific prompt $\bm{P}_s$ selected by the majority of samples might not be ideal. 
This is because the samples probably come from a target domain that has not been encountered before. 
As a result, none of the prompts in the memory buffer correspond to the domain of the data.
To overcome this issue, we compute the average matching score between all samples and $\bm{k}_s$ to assess its reliability:
\begin{equation}\label{eq:reliability}
{r_s} = \frac{1}{B}\sum\limits_{i = 1}^B {\mathop{\rm cosine}\nolimits} ( {\bm{q}_i,{\bm{k}_s}} ).
\end{equation}
Intuitively, if $\bm{k}_s$ exhibits a high matching degree with the current samples, then $\bm{P}_s$ can be considered a reliable prompt for the domain to which the samples belong.
Conversely, insufficient matching suggests that the current samples are likely from an unexplored target domain.
In this case, it becomes necessary to allocate a new prompt in the memory buffer for the domain.
Therefore, we introduce a predefined threshold $\eta$. 
If $r_s$ falls below $\eta$, the previously selected domain-specific prompt is discarded, and a new prompt is initialized in the memory buffer to serve as the selection for the current data:
\begin{equation}\label{eq:initialization}
\begin{array}{*{20}{c}}
{\bm{k}_s} \leftarrow \frac{1}{B}\sum\limits_{i = 1}^B {{\bm{q}_i}},{\bm{P}_s} \leftarrow \bm{Z},\\
\mathcal{M} \leftarrow \mathcal{M} \cup \left\{ {\left( {{\bm{k}_s},{\bm{P}_s}} \right)} \right\},
\end{array}\quad{\rm{if }}\ {r_s} < \eta,
\end{equation}
where $\bm{Z} \in {\mathbb{R}^{{L_p} \times {D_e}}}$ is a random matrix used for initialization.
Note that for the first batch of samples, we directly initialize a new prompt, as the memory is empty at this stage.

\subsection{Optimization Objective}
After determining the domain-specific prompt $\bm{P}_s$ suitable for the current batch of samples $\mathcal{B}$, we can further use $\mathcal{B}$ to optimize  $\bm{P}_s$.
Existing CTTA methods primarily rely on entropy minimization~\cite{wang2021tent} or a teacher-student pseudo-labeling scheme~\cite{wang2022continual,Doebler2023robust} for model adaptation in unlabeled target domains.
However, entropy minimization can lead to overly confident predictions, causing the model to collapse into a limited set of categories, while maintaining both a teacher and a student model requires additional computational resources.
Besides, the intrinsic structural information~\cite{xia2022maximum} of the target domain remains underexplored in previous studies.

In this study, we aim to maximize the mutual information~\cite{liang2022source,cui2024adversarial} of the model's predictions on target samples.
The loss function for maximizing the mutual information is defined as follows:
\begin{equation}\label{eq:mi}
{\mathcal{L}_{mi}} = \frac{1}{{{B}}}\sum\limits_{i = 1}^{{B}} {H\left( {{\bm{p}_i}} \right)} - H\left( {\frac{1}{{{B}}}\sum\limits_{i = 1}^{{B}} {{\bm{p}_i}} } \right),
\end{equation}
where $\bm{p}_i$ is the model's predicted probability distribution for the target sample $\bm{x}_i \in \mathcal{B}$, and $H\left( {\bm{p}_i} \right) =  - \sum\nolimits_{k = 1}^K {p_i^k\log } p_i^k$ denotes the entropy function with $p_i^k$ being the $k$-th element of $\bm{p}_i$.
The first term of Eq.~\eqref{eq:mi} is the average entropy of the model's predictions, and minimizing it prevents the model from producing ambiguous predictions for individual target samples. 
The second term is the negative entropy of the average output of the model, and minimizing it improves prediction diversity within a batch, thereby reducing the risk of the model's predictions collapsing into a few categories.
By combining the two terms, the mutual information loss ${\mathcal{L}_{mi}}$ contributes to both individual prediction certainty and overall prediction diversity for the target data.

% We also incorporate a structural regularization to encourage the model to make more smooth predictions in the target domain.
% Specifically, we enforce consistency between the model's predictions on interpolated target samples and the interpolated sample labels~\cite{zhang2018mixup}.
We also incorporate structural regularization to ensure consistency between the model's predictions on interpolated target samples and the interpolated sample labels~\cite{zhang2018mixup}.
Let $\hat{\bm{p}}_i$ denote the category distribution predicted by the model for $\bm{x}_i$ prior to batch optimization, i.e., by the model using the domain-specific prompt $\bm{P}_s$ before being updated.
The pseudo label $\hat{y}_i$ for $\bm{x}_i$ is generated based on $\hat{\bm{p}}_i$ if the prediction confidence exceeds a threshold $\phi$:
\begin{equation}\label{eq:pseudo_label}
\hat{y}_i = {\mathop {\arg\max }\limits_{k} \hat{p}_i^k },\ {\rm{if }}\ {\max\limits_{k}\hat{p}_i^k} > \phi.
\end{equation}
Then, each pair of target samples assigned with pseudo labels $\bm{x}_i, \bm{x}_j \in \mathcal{B}$ is randomly mixed, and their pseudo labels are correspondingly mixed:
\begin{equation}\label{eq:mixup}
\begin{array}{c}
{\bm{x}_{ij}} = \lambda {\bm{x}_i} + (1 - \lambda ){\bm{x}_j},\\
{\hat{y}_{ij}} = \lambda {\hat{y}_i} + (1 - \lambda ){\hat{y}_j},
\end{array}
\end{equation}
% where $\lambda  \sim {\mathop{\rm Beta}\nolimits} \left( {\alpha ,\alpha } \right)$ and $\alpha \in (0, \infty)$ is the hyperparameter of the Beta distribution.
where $\lambda  \sim {\mathop{\rm Beta}\nolimits} \left( {\alpha ,\alpha } \right)$ with $\alpha$ set to 1.0, ensuring that the interpolation coefficient $\lambda$ is uniformly distributed over the interval $\left [0,1 \right ] $.
Denote by $\bm{p}_{ij}$ the model's prediction on the blended sample ${\bm{x}_{ij}}$.
The interpolation consistency regularization can be formulated as the cross-entropy loss between $\bm{p}_{ij}$ and ${\hat{y}_{ij}}$:
\begin{equation}\label{eq:ce}
{\mathcal{L}_{ic}} =  - \frac{1}{B}\sum\limits_{{\bm{x}_i},{\bm{x}_j} \in {\mathcal{B}}} {\log \left( {{p}_{ij}^{{\hat{y}_{ij}}}} \right)}.
\end{equation}
Intuitively, $\mathcal{L}_{ic}$ compels the model to predict not only on the original target samples but also across the interpolated regions between them, thereby enhancing the model's generalization ability in the target domain.

Overall, the ultimate optimization objective of our PAINT method is as follows:
\begin{equation}\label{eq:objective}
\min {\mathcal{L}_{mi}} + \beta {\mathcal{L}_{ic}},
\end{equation}
where $\beta$ is a hyperparameter balancing mutual information loss and interpolation consistency loss.
% In addition to the domain-specific prompt $\bm{P}_s$, we fine-tune the first three transformer blocks of the pre-trained model, i.e., $\left\{ {{\mathcal{T}_l}} \right\}_{l=1}^3$, using the objective in Eq.~\eqref{eq:objective}.
In addition to the domain-specific prompt $\bm{P}_s$, we fine-tune the feature encoder $\mathcal{F}(\cdot)$ of the pre-trained model using the objective in Eq.~\eqref{eq:objective}.
To maintain a relatively low computational overhead, only the first three blocks of $\mathcal{F}(\cdot)$ are updated during training.
These shallow blocks capture the common knowledge shared by all target domains.
While this idea could also be implemented by inserting domain-invariant prompts into the model~\cite{gan2023decorate,wang2022dualprompt}, our experiments revealed that this option is less effective than fine-tuning the shallow encoder blocks.

\begin{algorithm}[!h]
	\caption{Pseudocode of model adaptation in PAINT.}
	\label{alg:training}
	\begin{algorithmic}[1]
		\renewcommand{\algorithmicrequire}{ \textbf{Input:}} %Use Input in the format of Algorithm
		\REQUIRE ~~\\
        Target samples arriving sequentially in batches, a pre-trained source model, and hyperparameters
		\renewcommand{\algorithmicrequire}{ \textbf{Initalization:}}
		\REQUIRE ~~\\
		A memory buffer  $\mathcal{M} = \varnothing$.
        \FOR{each batch of samples $\mathcal{B}= \{\bm{x}_1, \bm{x}_2, \ldots, \bm{x}_B\}$}
            \renewcommand{\arraystretch}{2}
            \STATE \textcolor[rgb]{0.5,0.5,0.5}{/* Prompt allocation */}
            \renewcommand{\arraystretch}{1}
            \IF{$\mathcal{M} == \varnothing$}
                \STATE Allocate a new prompt $(\bm{k}_s, \bm{P}_s)$ in $\mathcal{M}$; \textcolor[rgb]{0.5,0.5,0.5}{// Eq.~(5)}
            \ELSE
                \FOR{each sample $\bm{x}_i \in \mathcal{B}$}
                    \STATE Extract the query features $\bm{q}_i$ and retrieve the best-matching prompt $(\bm{k}_s^i, \bm{P}_s^i)$ for $\bm{x}_i$; \textcolor[rgb]{0.5,0.5,0.5}{// Eq.~(2)}
                \ENDFOR
                \STATE Select the optimal prompt $(\bm{k}_s, \bm{P}_s)$ for $\mathcal{B}$ via majority voting; \textcolor[rgb]{0.5,0.5,0.5}{// Eq.~(3)}
                \STATE Compute the reliability $r_s$ of $(\bm{k}_s, \bm{P}_s)$; \textcolor[rgb]{0.5,0.5,0.5}{// Eq.~(4)}
                \IF{${r_s} < \eta$}
                    \STATE Discard the selected prompt $(\bm{k}_s, \bm{P}_s)$;
                    \STATE Allocate a new prompt $(\bm{k}_s, \bm{P}_s)$ in $\mathcal{M}$; \textcolor[rgb]{0.5,0.5,0.5}{// Eq.~(5)}
                \ENDIF
            \ENDIF
            
            \STATE \textcolor[rgb]{0.5,0.5,0.5}{/* Prompt tuning */}
            \FOR{each sample $\bm{x}_i \in \mathcal{B}$}
                \STATE Predict the class distribution $\bm{p}_i$ for $\bm{x}_i$; \textcolor[rgb]{0.5,0.5,0.5}{// Eq.~(1)}
                \STATE Generate the pseudo label $\hat{y}_i$ for $\bm{x}_i$; \textcolor[rgb]{0.5,0.5,0.5}{// Eq.~(7)}
            \ENDFOR
            \STATE Compute the mutual information loss $\mathcal{L}_{mi}$ and the interpolation consistency loss $\mathcal{L}_{ic}$; \textcolor[rgb]{0.5,0.5,0.5}{// Eq.~(6) and Eq.~(9)}
            \STATE Optimize the prompt value $\bm{P}_s$ and the shallow blocks of $\mathcal{F}(\cdot)$ based on $\mathcal{L}_{mi}$ and $\mathcal{L}_{ic}$; \textcolor[rgb]{0.5,0.5,0.5}{// Eq.~(10)}
            \STATE Update the prompt key $\bm{k}_s$ using a moving average strategy; \textcolor[rgb]{0.5,0.5,0.5}{// Eq.~(11)}
		\ENDFOR	
	\end{algorithmic}
\end{algorithm}

\subsection{Prompt Key Updating}
If the prompt key $\bm{k}_s$ is not newly initialized as shown in Eq.~\eqref{eq:initialization}, we update $\bm{k}_s$ using a moving average approach, aligning it with the average query features of the corresponding target domain:
\begin{equation}\label{eq:moving_average}
{\bm{k}_s} \leftarrow \gamma {\bm{k}_s} + \frac{{\left( {1 - \gamma } \right)}}{B}\sum\limits_{i = 1}^B {{\bm{q}_i}},
\end{equation}
% where $\gamma$ is a smoothing coefficient hyperparameter.
where $\gamma$ is a smoothing coefficient fixed at 0.8 to prevent dramatic fluctuations in the moving average.
In this manner, $\bm{k}_s$ is consistently refined and eventually stabilizes to achieve an alignment with the target domain.

For clarity, we present the pseudocode of model adaptation in our PAINT method in Algorithm~\ref{alg:training}.
In particular, Lines 3-15 describe the prompt allocation process via the dynamic query mechanism, while Lines 17-23 summarize the steps of prompt tuning.

\section{Experiments}\label{sec:experiments}
In this section, we provide a detailed description of the experimental settings and present a series of experimental results to demonstrate the effectiveness of our PAINT method.

\subsection{Datasets}
We conducted experiments on three benchmark datasets for out-of-distribution evaluation, i.e., CIFAR10-C, ImageNet-C, and ImageNet-R.
%Specifically, CIFAR10-C and ImageNet-C~\cite{hendrycks2019robustness} contain corrupted images of CIFAR10 and ImageNet, respectively, with 15 types of corruptions across four categories, each type having five severity levels.
%ImageNet-R~\cite{hendrycks2021many} consists of various artistic renditions of images of object classes from ImageNet.

\textbf{CIFAR10-C and ImageNet-C}~\cite{hendrycks2019robustness} are corrupted versions of the original CIFAR10 and ImageNet datasets, respectively.
They are designed to evaluate model robustness against various common corruptions and perturbations. 
CIFAR10-C introduces a series of corruptions to the test set of CIFAR10 (10,000 images across 10 categories), while ImageNet-C applies these corruptions to the validation set of ImageNet (50,000 images across 1,000 categories).
Both datasets contain a total of 15 distinct corruption types, grouped into four main categories: noise, blur, weather, and digital.
Each corruption type is applied at five different severity levels, ranging from mild to severe.
This setup facilitates the study of how models perform under gradually worsening conditions.
% Figure~\ref{fig:example_c} shows images with varying corruption types and severity levels.

\begin{figure}[t]
    \centering
    \begin{minipage}{0.475\textwidth}
        \centering
	\subfloat[15 corruption types applied to images.]{\includegraphics[width=1.0\linewidth]{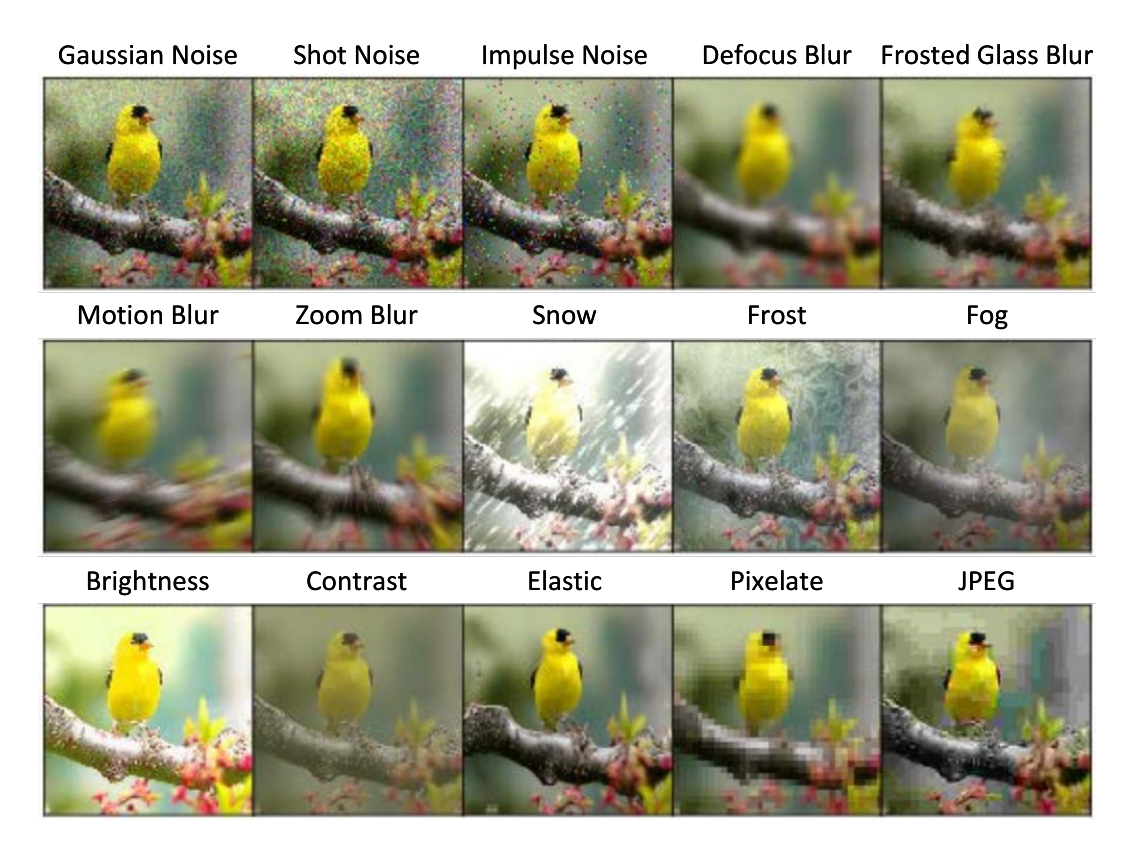}
		\label{fig:corruption}}
%\caption{Images with various corruption types and severity levels on CIFAR10-C and ImageNet-C. The images are sourced from~\cite{hendrycks2019robustness}.}        
    \end{minipage}
    \hspace{0.025\textwidth}
    \begin{minipage}{0.475\textwidth}
        \centering
    \subfloat[Corrupted images at five severity levels.]{\includegraphics[width=1.0\linewidth]{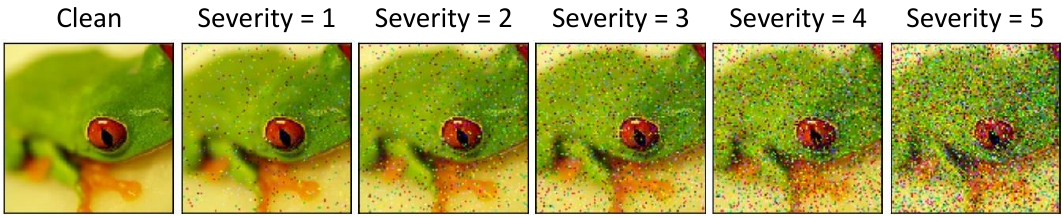}
		\label{fig:severity}}    
        
    \subfloat[Example images in different artistic renditions.]{\includegraphics[width=1.0\linewidth]{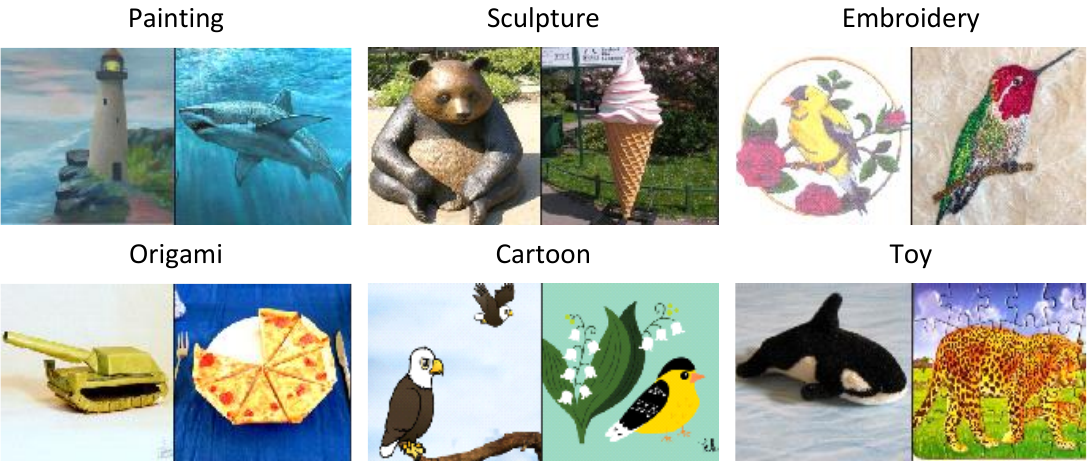}
		\label{fig:example_r}}
        
    \end{minipage}
\caption{Images with various corruption types and severity levels on CIFAR10-C and ImageNet-C, as well as some examples in different artistic renditions from ImageNet-R. The images are sourced from~\cite{hendrycks2019robustness} and~\cite{hendrycks2021many}, respectively.} 
\label{fig:dataset}
\end{figure}

\textbf{ImageNet-R}~\cite{hendrycks2021many} is a variant of the ImageNet dataset specifically crafted to evaluate the robustness of image classifiers to domain shifts caused by artistic renditions. 
Unlike the traditional corruptions in CIFAR10-C and ImageNet-C, ImageNet-R focuses on shifts that arise from different visual styles and representations.
The dataset comprises 30,000 images from 200 ImageNet classes, with objects depicted in various artistic forms, such as paintings, cartoons, graffiti, embroidery, and sculptures.
% Figure~\ref{fig:example_r} presents example images in different artistic renditions.
Figure~\ref{fig:dataset} shows the images with varying corruption types and severity levels on CIFAR10-C and ImageNet-C, as well as some examples in different artistic renditions from ImageNet-R.

\subsection{Evaluation Protocol}
Following previous work~\cite{wang2022continual}, we continuously adapted the pre-trained source model to the test samples from different target domains that arrived sequentially without resetting the model.
The model was updated with one gradient step for each batch of samples and then made predictions on them immediately.
On CIFAR10-C and ImageNet-C, each corruption type at its highest severity level corresponds to an individual target domain, while various rendition styles are regarded as target domains on ImageNet-R.
We employed the models trained on the original CIFAR10 and ImageNet as our pre-trained source models.
For each dataset, we evaluated the classification accuracy for each target domain, as well as the average accuracy across different domains.

\subsection{Baselines}
We compared PAINT with several recently proposed CTTA methods based on ResNet and ViT backbones.
The baseline methods include CoTTA~\cite{wang2022continual}, EATA~\cite{niu2022efficient}, ETA~\cite{niu2022efficient}, TENT~\cite{wang2021tent}, SANTA~\cite{Chakrabarty2023santa}, VDP~\cite{gan2023decorate}, CPL~\cite{goyal2022test}, RMT~\cite{Doebler2023robust}, and SAR~\cite{niu2023towards}.
A concise overview of these methods has been presented in Section~\ref{sec:related_work}.
For ResNet-based baselines, we directly cited the results from their original papers or other published papers~\cite{yu2024domain} if available.
As the results of all ViT-based baselines were not reported in existing literature,  we reproduced these methods, strictly adhering to the hyperparameters specified in their original papers.
% Conversely, we reproduced the results of all ViT-based baselines, as they were not reported in existing literature.

\begin{table*}[t]
\centering
\begin{adjustbox}{width=1.0\linewidth}
 	\begin{tabular}{l | ccc | cccc | cccc | cccc | >{\columncolor{blue!8}}c}
\toprule
 	& \multicolumn{3}{c|}{Noise} & \multicolumn{4}{c|}{Blur} & \multicolumn{4}{c|}{Weather} & \multicolumn{4}{c|}{Digital} & \\
 	Method & Gauss. & Shot & Impul. & Defoc. & Glass & Motion & Zoom & Snow & Frost & Fog & Brit. & Contr. & Elastic & Pixel & JPEG & Avg.\\
\midrule
        WideResNet-28 &27.70&34.30&27.10&53.10&45.70&65.20&58.00&74.90&58.70&74.00&90.70&53.30&73.40&41.50&69.70&56.50 \\ 
        ~~$\bullet~$CoTTA &75.70&78.70&73.40&88.40&72.40&87.80&89.70&85.20&85.90&87.60&92.50&89.40&81.70&86.60&82.70&83.80 \\
        ~~$\bullet~$EATA &75.70&80.70&72.40&87.40&71.40&85.60&88.00&84.10&85.40&84.60&90.40&86.70&79.40&83.70&78.20&82.20 \\
        ~~$\bullet~$TENT &75.20&79.40&71.40&85.60&68.90&83.50&85.90&80.90&81.40&81.40&87.80&79.70&74.30&79.20&75.10&79.30\\ 
        ~~$\bullet~$VDP &77.40&80.30&71.90&92.90&71.60&90.50&93.70&89.80&88.50&91.00&\textbf{98.50}&94.40&81.50&87.20&81.50&86.00\\ 
        ~~$\bullet~$SANTA &76.10&79.90&72.00&88.40&72.60&87.40&89.80&85.90&86.80&87.80&92.60&89.70&80.90&86.70&81.50&83.90\\
        ~~$\bullet~$RMT &75.50&80.00&74.50&86.10&75.40&85.10&86.70&84.00&84.20&84.40&88.90&85.00&81.70&85.40&83.10&82.70\\
\midrule

ViT-B/16  
&77.10 &79.38 & 81.75 &88.06 &67.58 &84.12 &89.58 &92.41 &90.54 &81.98 &96.47 &65.56 &79.83 &63.46 &80.50&81.22\\
~~$\bullet~$CoTTA 
& 77.10 & 79.39 & 81.74&88.07&67.56&84.12&89.58&92.41&90.57&81.97&96.48&65.58&79.90&63.48&80.49 &81.23 \\ 
~~$\bullet~$ETA  & 78.09 & 80.26 &
    82.28&88.26&67.63&84.48& 89.69&92.43&90.66&82.17&96.48&66.52&80.12&62.88&80.69 &81.51 \\ 
~~$\bullet~$CPL  
&83.52 &88.25 &89.20 &91.47& 78.55&90.67 &94.08& 94.37&94.66& 91.61&97.21&93.11&87.45&88.01&88.01&90.01\\
~~$\bullet~$RMT &80.97&85.88&86.68&90.57&\textbf{85.71}&90.07&93.83&93.84&94.47&91.67&96.31&91.95&\textbf{89.80}&91.62&\textbf{89.93}&90.35 \\
~~$\bullet~$SAR & 85.46 & 85.74 & 88.64 & 90.79 & 80.57 & 89.10 & 90.90 & 92.40 & 91.11 & 88.92 & 96.48 & 86.88 & 85.27 & 76.73 & 85.63 &87.64   \\
\rowcolor{orange!20}~~$\bullet~$PAINT&\textbf{86.26} &\textbf{89.08} &\textbf{90.56} &\textbf{93.48} &84.21 &\textbf{92.75} &\textbf{95.20} &\textbf{94.55} &\textbf{95.23} &\textbf{94.43} &96.85 &\textbf{95.90 }&89.14&\textbf{92.94} &89.43 &\textbf{92.00}\\
\bottomrule
	\end{tabular}
	\end{adjustbox}
\caption{Classification accuracy on CIFAR10-C at the highest corruption severity level. Each column represents the performance on a target domain, with data from different target domains arriving sequentially from left to right.}
\label{tab:cifar-c}
\end{table*}

\subsection{Implementation Details}
We followed the experimental setups described in previous CTTA studies~\cite{wang2022continual, niu2022efficient}.
All input images were resized to $224 \times 224$ and normalized to the range $[0, 1]$.
The random seed was fixed at 0 for all methods during training.
All experiments were conducted in PyTorch using a single NVIDIA GeForce RTX 3090 GPU and 128GB system memory.

For PAINT, we employed the pre-trained ViT-B/16 source model for prompt tuning, where the dimensions of both the patch embeddings and visual features are 768.
During adaptation, the batch size and learning rate were configured as 50 and 0.05, respectively.
We set the prompt length $L_p$ to 2, consistent with previous works~\cite{khattak2023maple}.
The matching score threshold $\eta$ and prediction confidence threshold $\phi$ were set to 0.2 and 0.6, respectively.
For the loss balancing hyperparameter $\beta$, we chose $\beta=1$.
The impact of various hyperparameters on model performance will be discussed later.
For the sake of reproducibility, the code of our PAINT method has been released at \url{https://github.com/Cadezzyr/PAINT}.

\begin{table*}[t]
\centering
\begin{adjustbox}{width=1.0\linewidth}
 	\begin{tabular}{l | ccc | cccc | cccc | cccc | >{\columncolor{blue!8}}c}
\toprule
 	& \multicolumn{3}{c|}{Noise} & \multicolumn{4}{c|}{Blur} & \multicolumn{4}{c|}{Weather} & \multicolumn{4}{c|}{Digital} & \\
 	Method & Gauss. & Shot & Impul. & Defoc. & Glass & Motion & Zoom & Snow & Frost & Fog & Brit. & Contr. & Elastic & Pixel & JPEG & Avg.\\
\midrule
        ResNet-50 & 2.20  & 2.90  & 1.80  & 17.80  & 9.80  & 14.50  & 22.50  & 16.80  & 23.40  & 24.60  & 59.00  & 5.50  & 17.10  & 20.70  & 31.60  & 18.00  \\ 
        ~~$\bullet~$CoTTA &15.30&17.90&19.40&18.70&21.00&31.40&42.50&39.70&39.50&51.70&63.40&33.90&52.80&58.80&54.00&37.30 \\
        ~~$\bullet~$EATA     &35.00 &38.10 &36.80 &33.80 &34.20 &47.30 &53.20 &51.10 &45.60 &59.70 &68.00 &44.20 &57.20 &60.40 &54.70 &47.95  \\
        ~~$\bullet~$TENT &18.40&25.40&27.30&22.40&26.20&34.50&44.70&38.40&37.00&48.30&61.80&27.90&49.20&52.60&46.70&37.40 \\
        ~~$\bullet~$SANTA &25.90&27.10&28.40&24.30&25.90&35.80&44.50&44.40&37.10&53.40&63.90&30.10&49.40&55.70&51.50&39.90  \\ 
        ~~$\bullet~$RMT &19.80&23.60&25.50&22.90&25.60&33.80&42.40&43.00&40.90&52.00&60.90&39.40&52.70&57.50&56.60&39.80\\
\midrule
        ViT-B/16  &46.90&45.80 &
    47.00 &31.35 &21.30 &42.30 &38.20 &49.55 &45.40 &43.45 &74.45 &9.00 &41.85 &62.15 &64.65 &44.22 \\
~~$\bullet~$CoTTA & 50.48 & 56.42 & 58.32&39.48&42.50&52.02&46.76&56.92&59.84&53.12&70.98&15.74&\textbf{62.58}&68.26&67.54 &53.40 \\ 
~~$\bullet~$ETA  &56.02&60.40&58.60
    &46.96&46.40&53.48&53.94&56.78&59.44&58.00&74.52&37.26&55.76&64.88&65.96 &56.56  \\
~~$\bullet~$CPL  &51.30 &57.60 &57.46 &44.18 &42.02 &54.16 &51.06 &60.88 &58.98 &59.58 &76.10 &47.28 &28.68 &47.06 &2.48 &49.25\\
~~$\bullet~$RMT &49.28&52.70&55.28&40.38&40.98&51.54&48.30&56.64&56.72&53.70&69.66&26.64&62.46&68.56&68.20&53.41 \\
~~$\bullet~$SAR &\textbf{56.22}&\textbf{60.48}&\textbf{59.30}   &48.64&50.06&55.46&54.50&59.88&60.82&60.22&75.94&45.04&57.76&68.20&68.40&58.73   \\
\rowcolor{orange!20}~~$\bullet~$PAINT&54.34&58.16&58.14   &\textbf{50.18}&\textbf{50.54}&\textbf{57.62}&\textbf{55.66}&\textbf{62.46}&\textbf{63.00}&\textbf{65.92}&\textbf{77.58}&\textbf{53.30}&56.22&\textbf{71.94}&\textbf{69.66}&\textbf{60.31} \\
\bottomrule
	\end{tabular}
	\end{adjustbox}
\caption{Classification accuracy on ImageNet-C at the highest corruption severity level.   }
\label{tab:imagenet-c}
\end{table*}

\begin{table*}[t]
\centering
\begin{adjustbox}{width=1.0\linewidth}
 	\begin{tabular}{l | ccc | cccc | cccc | cccc | >{\columncolor{blue!8}}c}
\toprule
 	% & \multicolumn{3}{c|}{Noise} & \multicolumn{4}{c|}{Blur} & \multicolumn{4}{c|}{Weather} & \multicolumn{4}{c|}{Digital} & \\
 	Method & Art & Cart. & Dev. & Emb.& Graf. & Graphic & Misc & Orig. & Paint & Sculp. & Sketch & Stick. & Tat. & Toy & Video & Avg.\\
\midrule
ViT-B/16 &50.72 &25.17 &57.33
    &31.86 &32.09 &43.17 &48.18 &34.18 &68.04 &44.15 &35.50 &34.06 &29.89 &42.62 &53.38 &42.02\\
~~$\bullet~$CoTTA 
&51.31&27.99&61.26
    &34.49&35.39&50.47&55.74&40.91&74.85&54.52&56.62&39.37&39.29&51.62&63.92 &49.18 \\ 
~~$\bullet~$ETA  &53.53&37.63&70.37&45.15&46.12&\textbf{60.56}&63.46&53.45&\textbf{78.98}&60.70&\textbf{67.65}&\textbf{47.64}&\textbf{50.44}&55.13&\textbf{64.56}&57.02 \\
~~$\bullet~$CPL  
&37.17&14.20&43.09&19.94&19.98&29.81&35.80&20.73&53.90&30.77&25.10&18.70&20.08&28.39&30.87&28.57\\ 
~~$\bullet~$RMT &52.29&29.09&63.15&34.07&33.90&50.47&56.13&38.91&73.97&52.76&55.78&39.76&41.55&52.02&63.50&49.16 \\
~~$\bullet~$SAR &51.05&26.22&59.08   &32.27&33.16&45.65&51.21&35.45&70.62&46.99&44.89&37.20&33.90&46.13&57.17&44.73  \\
\rowcolor{orange!20}~~$\bullet~$PAINT&\textbf{57.33}&\textbf{42.70}&\textbf{71.94}    &\textbf{54.02}&\textbf{51.33}&58.07&\textbf{64.46}&\textbf{60.91}&76.58&\textbf{61.20}&65.84&45.67&50.39&\textbf{55.20}&62.38&\textbf{58.53}\\
\bottomrule
	\end{tabular}
	\end{adjustbox}
\caption{Classification accuracy on ImageNet-R.}
\label{tab:imagenet-r}
\end{table*}

\subsection{Performance Comparison}
\textbf{Results on CIFAR10-C.}
As shown in Table~\ref{tab:cifar-c}, the ViT-based source model is much better than its ResNet-based counterpart on CIFAR10-C, demonstrating the superior transferability of ViT.
However, the source models exhibit relatively low accuracy and fall behind nearly all CTTA methods, indicating the necessity for model adaptation during testing.
Our PAINT method achieves the highest performance in 11 out of the 15 target domains.
It exceeds the source model and the second-best CTTA method RMT in average accuracy by 10.78\% and 1.65\%, respectively.

\textbf{Results on ImageNet-C.}
As illustrated in Table~\ref{tab:imagenet-c}, all CTTA methods demonstrate increasingly pronounced advantages over the source models on ImageNet-C.
PAINT not only outperforms the other methods in average accuracy, but also emerges as the sole method to achieve a result exceeding 60\%.
Notably, when tackling the adaptation task in the ``Contr.'' domain, the ViT-based source model achieves an accuracy of only 9.00\%.
In contrast, PAINT shows significant superiority with an improvement of at least 6.02\%, indicating its greater proficiency in addressing the most challenging tasks compared to the other methods.

\textbf{Results on ImageNet-R.}
Table~\ref{tab:imagenet-r} presents the results of all ViT-based methods on ImageNet-R.
As expected, PAINT once again obtains the best performance, improving the average accuracy of the source model from 42.02\% to 58.53\%.
It beats the runner-up method ETA by 1.51\%.
These results confirm the effectiveness of PAINT in mitigating continuous domain shifts for CTTA.

\begin{table}[t]
	\centering
    \begin{adjustbox}{width=0.975\linewidth}
	\begin{tabular}{c c c c c c c c}
		\toprule
	    & Source & CoTTA & ETA & CPL & RMT  & SAR & PAINT\\
		\midrule
		Accuracy & 91.31 $\pm$ 0.00 & 91.35 $\pm$ 0.19 & 91.39 $\pm$ 0.17 & 95.59 $\pm$ 0.12 & 95.67 $\pm$ 0.40 & 92.96 $\pm$ 0.02 & \textbf{96.67} \bm{$\pm$} \textbf{0.14}\\
		\bottomrule
	\end{tabular}
    \end{adjustbox}
\caption{Performance comparison between different methods on CIFAR10C in the gradually changing scenario. We report the mean and standard deviation of average accuracy over ten shuffled sequences of target domains.}
	\label{tab:gradual}
\end{table}

\subsection{Comparison in Gradually Changing Scenario}
In CTTA, some domain shifts may gradually evolve over time, leading to the gradually changing scenario~\cite{Doebler2023robust}.
Following prior work~\cite{wang2022continual}, we simulate this scenario on CIFAR10-C by constructing sequences of target domains with progressively changing severity across the 15 corruption types:
{\scriptsize
$\underbrace { \ldots 2 \to 1}_{t - 1\ {\rm{and\ before}}} \xrightarrow[\rm{change}]{\rm{type}} \underbrace {1 \to 2 \to 3 \to 4 \to 5 \to 4 \to 3 \to 2 \to 1}_{{\rm{corruption\ type\ }}t{\rm{,\ gradually\ changing\ severity}}} \xrightarrow[\rm{change}]{\rm{type}} \underbrace {1 \to 2 \to  \ldots }_{t + 1{\rm{\ and\ on}}}$
}, where the severity within each corruption type $t$ increases from the lowest (1) to the highest (5) before returning to the lowest.
On CIFAR10-C, each corruption type is regarded as an individual target domain.
% We randomly shuffled the order of target domains 10 times and calculated the mean and standard deviation of average accuracy over the ten shuffled sequences.
We performed ten independent random shuffles of the target domain sequence and computed the mean and standard deviation of the average accuracy across these shuffled sequences.
As demonstrated in Table~\ref{tab:gradual}, our PAINT method still surpasses the other methods in the gradually changing scenario, achieving a mean average accuracy of 96.67\% with a relatively low standard deviation of 0.14\%.
This corresponds to a 5.36\% and 5.32\% improvement in the mean value compared to the source model and CoTTA, respectively.
Notably, compared to adapting to domains with only the highest severity level (see Table~\ref{tab:cifar-c}), adapting to target domains with lower severity levels consistently yields better results for all methods.

\begin{figure}[t]
	\centering
    \subfloat[Classification accuracy on the source domain on CIFAR10-C.]{\includegraphics[width=0.6\linewidth]{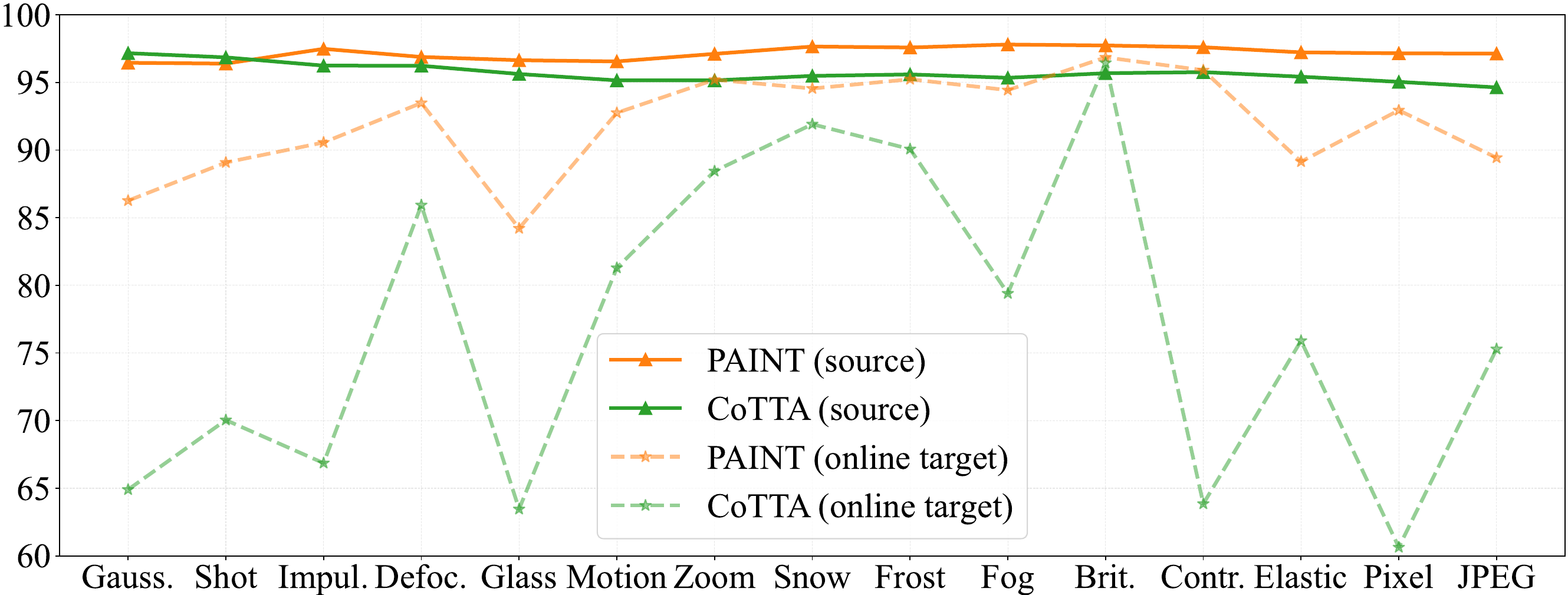}
		\label{fig:antiforget_source_cifar10}}

	\subfloat[Classification accuracy on the source domain on ImageNet-C.]{\includegraphics[width=0.6\linewidth]{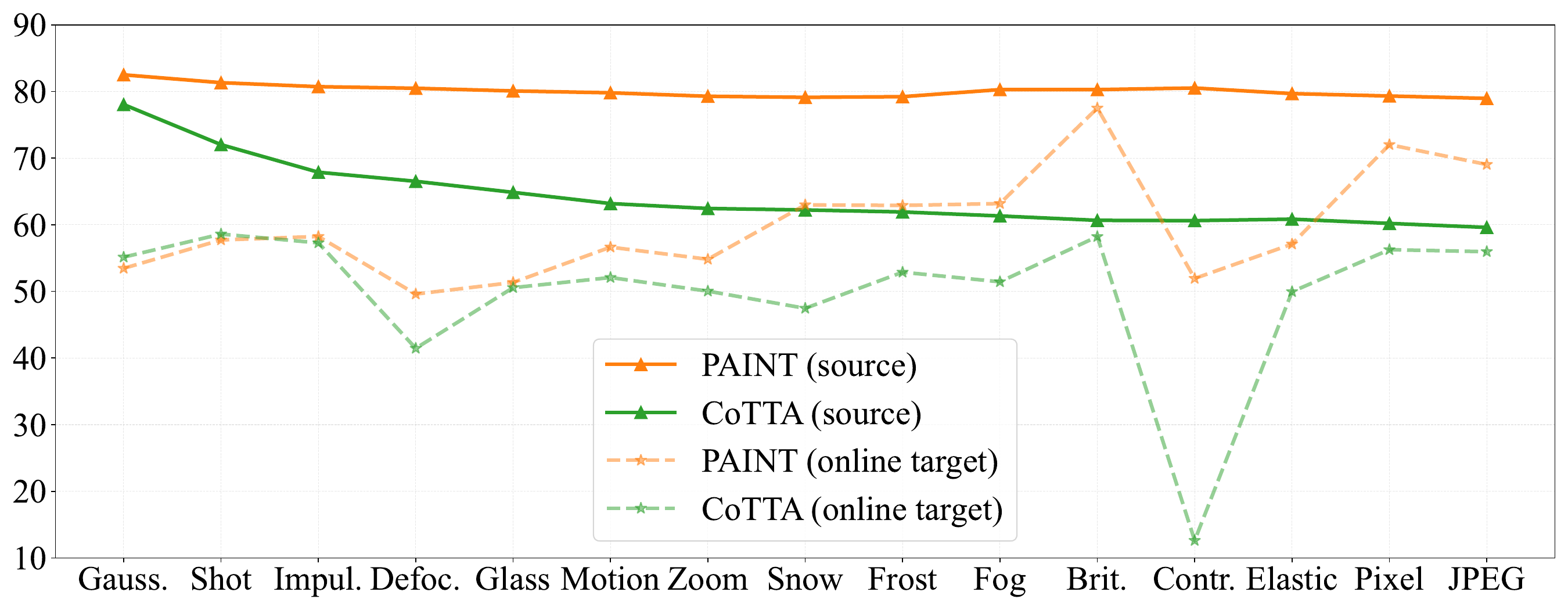}
		\label{fig:antiforget_source_c}}

	\subfloat[Classification accuracy on the source domain on ImageNet-R.]{\includegraphics[width=0.6\linewidth]{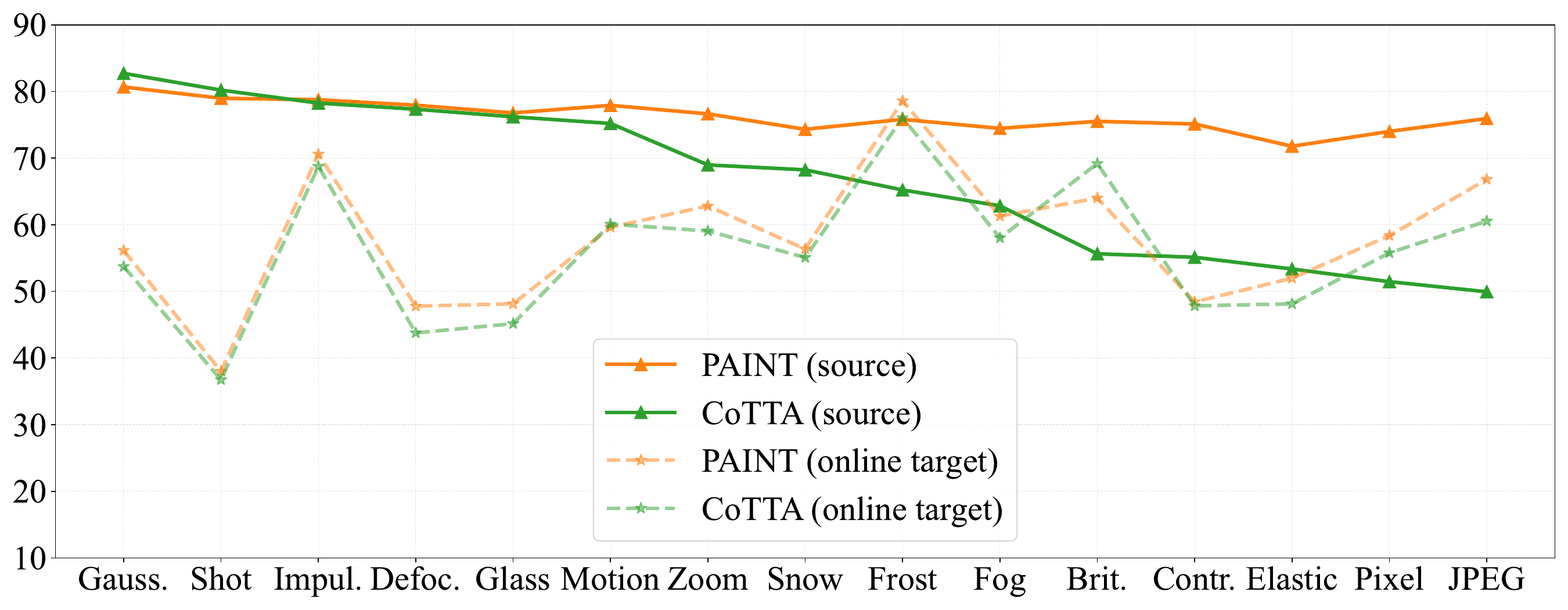}
		\label{fig:antiforget_source_r}}
	\caption{Anti-forgetting capabilities of PAINT and CoTTA.
% We report the average accuracy (\%) for the tasks with the same target domain on Office-Home, respectively.
}
	\label{fig:antiforget}
\end{figure}

\subsection{Anti-Forgetting Capability Test}
Following previous studies~\cite{niu2022efficient}, we examined the anti-forgetting capability of our PAINT method by monitoring the model's performance changes on the original source data during test-time adaptation.
In Figure~\ref{fig:antiforget_source_cifar10}, we compare the accuracy of PAINT with that of CoTTA on the source data of CIFAR10-C.
As can be seen, although both models maintain over 95\% accuracy on the clean source data, CoTTA's accuracy slightly declines over time, while PAINT exhibits more stable performance.
% In Figure~\ref{fig:antiforget_source_c}, we compare the accuracy of PAINT with that of CoTTA on the source data of ImageNet-C.
Figure~\ref{fig:antiforget_source_c} illustrates their comparison results on ImageNet-C.
PAINT consistently outperforms CoTTA in every online target domain as well as the source domain during adaptation.
Meanwhile, it is clear that CoTTA suffers from significant catastrophic forgetting, with its accuracy on the source data gradually decreasing from 78.04\% to 59.60\% over time.
Comparatively, PAINT exhibits more stable performance in the source domain, with accuracy consistently maintaining around 80\%.
Figure~\ref{fig:antiforget_source_r} presents the comparison results on ImageNet-R.
A similar trend can be observed, where CoTTA is even slightly ahead of PAINT at the beginning; however, as model adaptation progresses, its accuracy in the source domain declines rapidly and eventually lags behind PAINT by as much as 26 percentage points.
Note that throughout testing, PAINT seeks to allocate a distinct domain-specific prompt to all source data via the dynamic query mechanism.
In this way, the source domain and target domains can be learned independently, thereby mitigating the inter-domain interference.
The above results verify that PAINT is more effective at preserving source knowledge and demonstrates a stronger ability to prevent catastrophic forgetting.

\subsection{Ablation Study}

%\begin{table}[t]
%	\centering
%    %\begin{adjustbox}{width=0.95\linewidth}
%	\begin{tabular}{c c c c c c}
%		\toprule
%		$\mathcal{L}_{ent}$ & $\mathcal{L}_{mi}$ & $\mathcal{L}_{ic}$ & $\bm{P}_s$ & $\mathcal{F}(\cdot)$  & {Accuracy}\\
%		\midrule
%		\checkmark  & \ding{53} & \ding{53} & \checkmark & \checkmark & 47.45\\
%        \ding{53}  & \checkmark & \ding{53} & \checkmark & \checkmark & 52.97 \\
%        \ding{53}  & \checkmark & \checkmark & \checkmark & \checkmark  & \textbf{60.31} \\
%        \ding{53}  & \checkmark & \checkmark & \ding{53} & \checkmark & 58.58 \\
%        \ding{53}  & \checkmark & \checkmark & \checkmark & \ding{53} & 59.46 \\
%		\bottomrule
%	\end{tabular}
%    %\end{adjustbox}
%\caption{Contribution of different components of PAINT.}
%	\label{tab:ablation}
%\end{table}

Table~\ref{tab:ablation} lists the results of several variants of our PAINT method on ImageNet-C when different key components are omitted.
% We can see that using mutual information loss $\mathcal{L}_{mi}$ achieves better performance than entropy minimization loss $\mathcal{L}_{ent}$, and this advantage is significantly amplified by further introducing interpolation consistency loss $\mathcal{L}_{ic}$.
In PAINT, we perform test-time model adaptation by jointly optimizing the mutual information loss $\mathcal{L}_{mi}$ and the interpolation consistency loss $\mathcal{L}_{ic}$.
We can see that optimizing $\mathcal{L}_{mi}$ alone outperforms the traditional entropy minimization loss $\mathcal{L}_{ent}$~\cite{wang2021tent}, with performance gains that are further magnified when $\mathcal{L}_{ic}$ is incorporated.
These observations validate the rationality of our choice of optimization objective in PAINT.
As previously mentioned, PAINT simultaneously updates the domain-specific prompt $\bm{P}_s$ and the domain-agnostic shallow blocks of feature encoder $\mathcal{F}(\cdot)$ during adaptation.
The variants that do not include either $\bm{P}_s$ or $\mathcal{F}(\cdot)$ yield inferior results compared to PAINT, leading to a 1.73\% and 0.85\% drop in average accuracy, respectively.
This finding indicates that learning both domain-specific and domain-agnostic knowledge are crucial for enhancing model performance.

\begin{figure}[t]
    \centering
    \begin{minipage}{0.45\textwidth}
	\centering
	\begin{tabular}{c c c c c c}
		\toprule
		$\mathcal{L}_{ent}$ & $\mathcal{L}_{mi}$ & $\mathcal{L}_{ic}$ & $\bm{P}_s$ & $\mathcal{F}(\cdot)$  & {Accuracy}\\
		\midrule
		\checkmark  & \ding{53} & \ding{53} & \checkmark & \checkmark & 47.45\\
        \ding{53}  & \checkmark & \ding{53} & \checkmark & \checkmark & 52.97 \\
        \ding{53}  & \checkmark & \checkmark & \checkmark & \checkmark  & \textbf{60.31} \\
        \ding{53}  & \checkmark & \checkmark & \ding{53} & \checkmark & 58.58 \\
        \ding{53}  & \checkmark & \checkmark & \checkmark & \ding{53} & 59.46 \\
		\bottomrule
	\end{tabular}
	\captionof{table}{Contribution of different components of PAINT.}
	\label{tab:ablation} 
    \end{minipage}
    \hspace{0.025\textwidth}
    \begin{minipage}{0.5\textwidth}
        \centering
    \begin{tabular}{c | c c c c c}
		\toprule
		$\beta$ & 0.01 & 0.1 & 1 & 10  & 100 \\
		\midrule
		Accuracy  & 54.96 & 58.25 & \textbf{60.31} & 59.61 & 58.69\\
		\bottomrule
	\end{tabular}
    %\end{adjustbox}
    \captionof{table}{Effect of the balancing hyperparameter $\beta$.}
% \caption{Effect of the balancing hyperparameter $\beta$.}
	\label{tab:beta}
        
    \end{minipage}
\end{figure}

\begin{figure}[t]
	\centering
	\subfloat[Prompt length]{\includegraphics[width=0.3\linewidth]{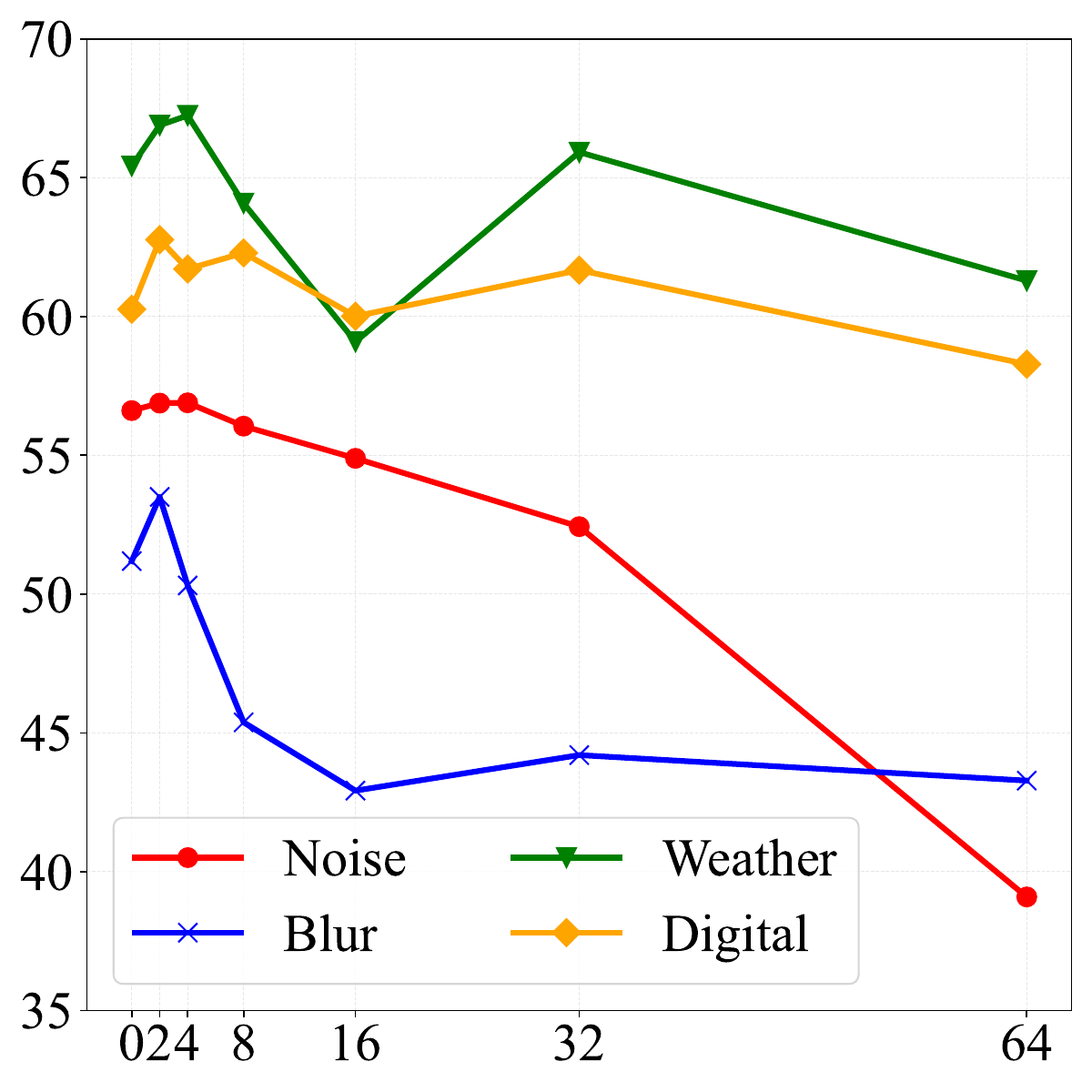}
		\label{fig:prompt_lepth}}
    \hspace{0.05\textwidth}
	\subfloat[Threshold]{\includegraphics[width=0.3\linewidth]{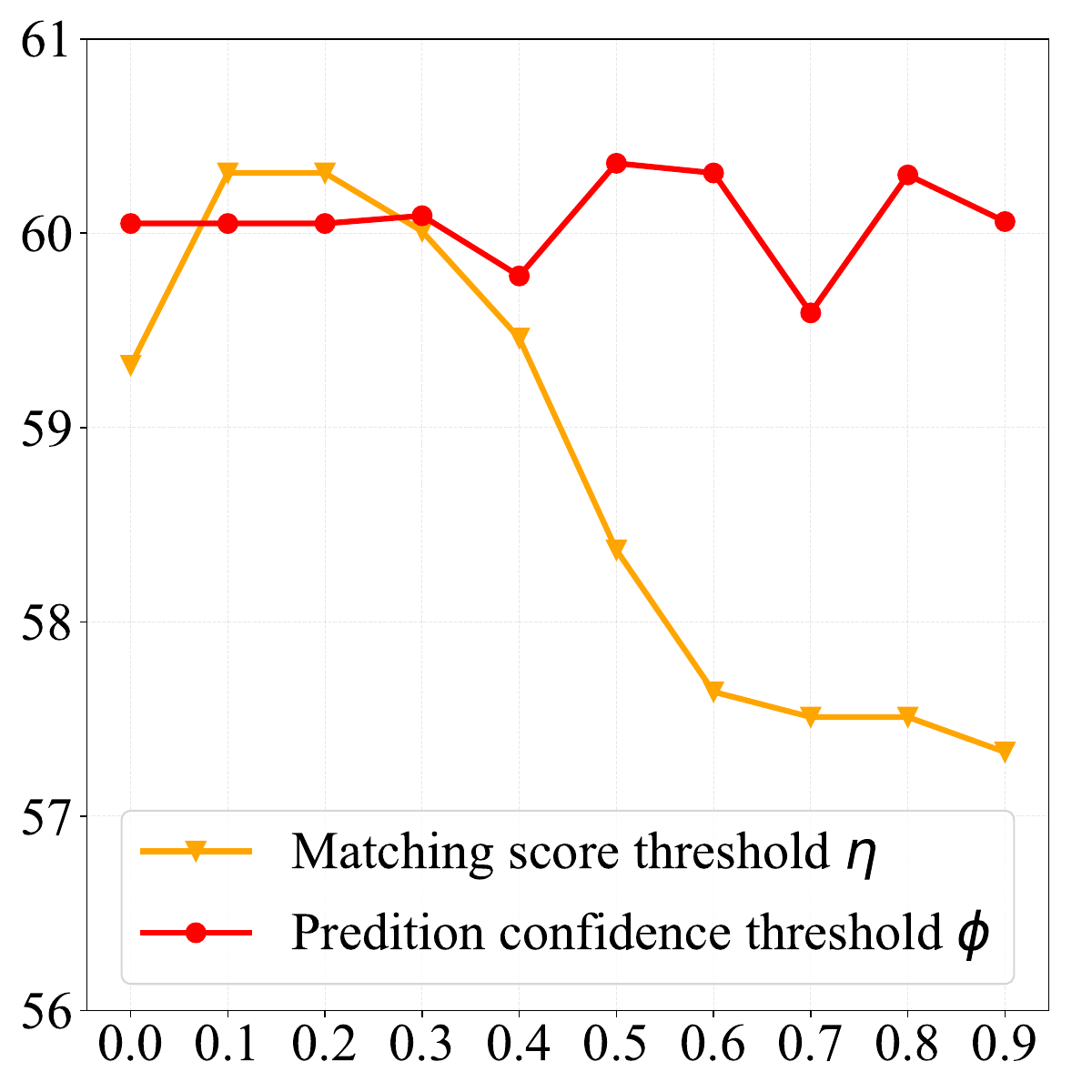}
		\label{fig:threshold}}
	\caption{Effects of the prompt length and thresholds.
% We report the average accuracy (\%) for the tasks with the same target domain on Office-Home, respectively.
}
	\label{fig:hyperparameter}
\end{figure}

\subsection{Hyperparameter Analysis}
We evaluate the influence of different hyperparameters on PAINT's performance on ImageNet-C:

\textbf{Effect of Prompt Length.}
Figure~\ref{fig:prompt_lepth} shows the impact of prompt length across four corruption categories in target domains:
We notice that the best results are achieved with short prompts (e.g., two or four prompt tokens), while further extending the length severely degrades model performance.
One possible explanation is that longer prompts introduce additional model parameters.
In CTTA, the model must make predictions immediately after being updated with a small amount of data in a batch.
The increase in model parameters may lead to overfitting, particularly in an environment where the distribution of test data is continuously evolving.

\textbf{Effect of Threshold.}
Figure~\ref{fig:threshold} depicts how the threshold hyperparameters $\eta$ and $\phi$ affect PAINT's performance.
In Eq.~\eqref{eq:initialization}, $\eta$ represents the threshold for query and key matching scores in prompt allocation.
It is observed that the performance goes down gradually when $\eta$ exceeds 0.2.
A higher value of $\eta$ increases the likelihood of allocating a new prompt to each batch of samples.
This can result in samples from the same target domain failing to share a consistent domain-specific prompt, thereby reducing the model's accuracy.
In Eq.~\eqref{eq:pseudo_label}, $\phi$ denotes the confidence threshold for the pseudo labels of target samples.
The best results are obtained when $\phi=0.5$ or $\phi=0.6$.
Overall, the performance of PAINT remains relatively stable as $\phi$ varies.

\textbf{Effect of Loss Weight.}
In Eq.~\eqref{eq:objective}, the hyperparameter $\beta$ controls the trade-off between the mutual information loss and the interpolation consistency loss.
Table~\ref{tab:beta} displays the results obtained by PAINT with different values of $\beta$.
We adjust $\beta$ over several orders of magnitude ranging from 0.01 to 100.
The performance improves progressively as $\beta$ changes from 0.01 to 1, but deteriorates with any further increase.

\begin{table}[t]
	\centering
    %\begin{adjustbox}{width=0.95\linewidth}
	\begin{tabular}{l c c c}
		\toprule
	    & Prompt number & Matching accuracy & Classification accuracy\\
		\midrule
		PAINT-Query & 11 & 53.33 & 60.31\\
        PAINT-Oracle & 15 & 100 & 60.55 \\
		\bottomrule
	\end{tabular}
    %\end{adjustbox}
\caption{Comparison between PAINT with our query mechanism (i.e., PAINT-Query) and a variant with perfect knowledge of the domain identity of target samples at test time (i.e., PAINT-Oracle) on ImageNet-C.}
	\label{tab:matching}
\end{table}

\subsection{Relationship between Query Accuracy and Model Performance}
As described in Section~\ref{sec:prompt_allocation}, our PAINT method utilizes a query mechanism that dynamically determine whether to select an existing domain-specific prompt or to allocate a new prompt for each batch of target samples.
To investigate the relationship between query accuracy and model performance, we compare PAINT against a variant in which the domain identity of target samples is provided at test time on ImageNet-C.
The comparison results are presented in Table~\ref{tab:matching}.
We observe that in PAINT's query mechanism, the matching accuracy between target samples and their true corresponding domain-specific prompts is not very high, at only 53.33\%. 
For the 15 target domains on ImageNet-C, PAINT allocates only 11 prompts throughout the entire adaptation process. 
Remarkably, despite this mismatch, PAINT remains robust, with its performance using the query mechanism only being slightly behind that of the variant where the domain identity of target samples is known (60.31\% versus 60.55\%).
We suspect that this robustness stems from the strong similarities between different target domains. 
For example, as illustrated in Table~\ref{tab:imagenet-c}, multiple target domains often belong to the same corruption category. 
Since the matching is performed based on the visual features of target samples, PAINT can implicitly leverage domain similarities.
Even if a mismatch occurs, PAINT may still select a prompt from a visually similar target domain, enabling it to achieve reasonable performance.

\begin{figure}[t]
	\centering
	\includegraphics[width=0.65\textwidth]{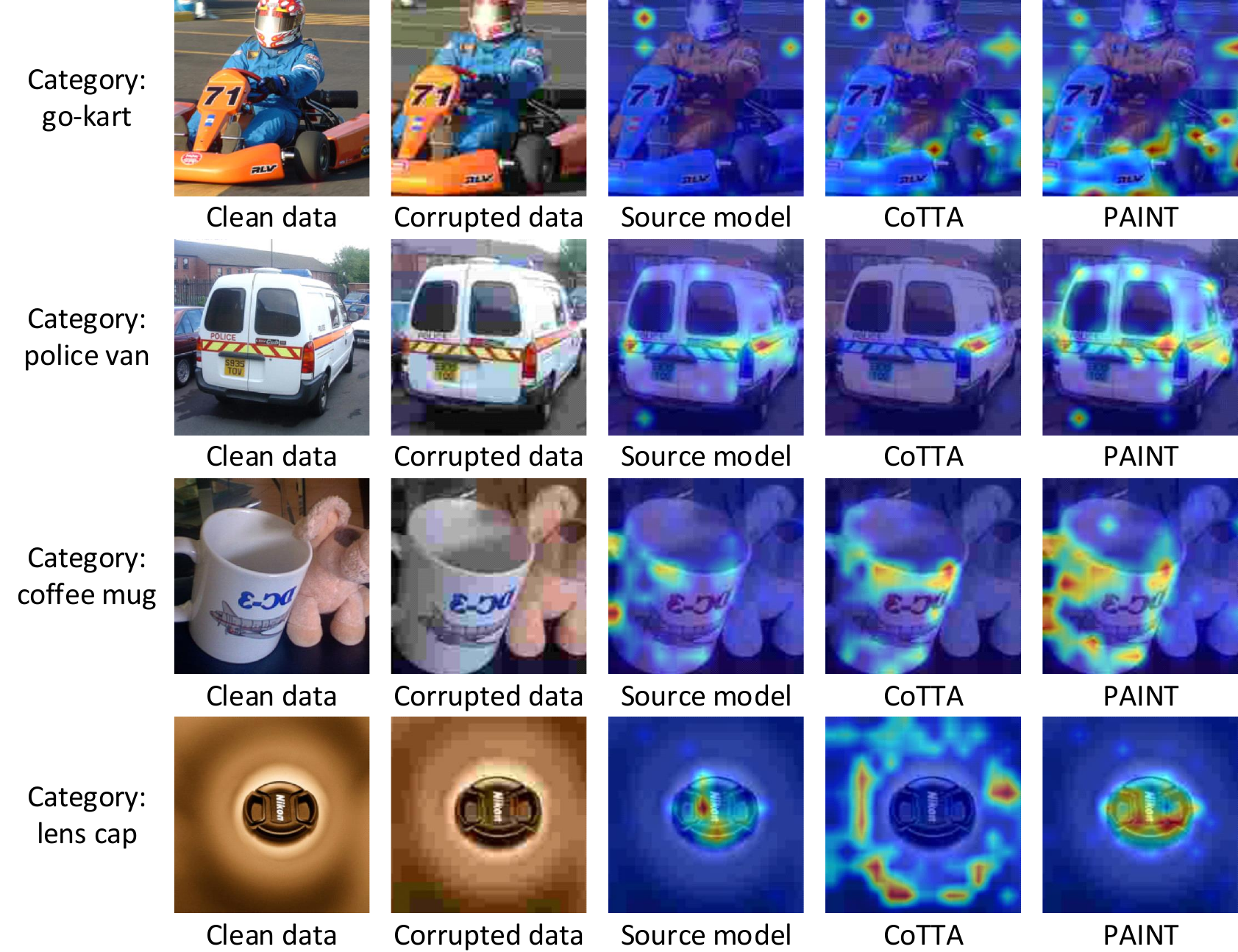}
\caption{Visualization of attention maps learned by the source model, CoTTA, and PAINT for some corrupted images in target domains. Each image is annotated with its category label and paired with its corresponding clean image in the source domain.}
\label{fig:attention_map}
\end{figure}

\subsection{Visualization}
% To gain deeper insight into the superiority of our PAINT method, we present visualizations of the attention maps learned by the source model, CoTTA, and PAINT for several corrupted images in target domains on ImageNet-C.
% Figure~\ref{fig:attention_map} illustrates these visualizations.
% The attention maps are generated by averaging the attention scores between the class token and the other patch tokens across all heads in the last transformer block of ViT~\cite{Dosovitskiy2021image}.
To gain deeper insight into the superiority of our PAINT method, we visualize the class activation maps (CAMs)~\cite{zhou2016learning} generated by the source model, CoTTA, and PAINT for several corrupted images in target domains on ImageNet-C.
To generate these CAMs, we first compute the model's output logits and use their gradients to weight the attention maps of each encoder block. 
The final CAMs are derived by averaging the weighted attention maps across all blocks.
Figure~\ref{fig:attention_map} presents these visualizations, where each image is annotated with its category label and paired with its corresponding clean image in the source domain.
It can be observed that PAINT focuses more accurately on the object of interest compared to the source model and CoTTA, effectively minimizing distractions from the background or unrelated objects.
In contrast, CoTTA occasionally fails to adapt the source model.  
For example, for the image of the lens cap category, CoTTA diverts the source model's attention to incorrect regions, whereas PAINT further refines the focus on important visual areas.

\section{Conclusion}\label{sec:conclusions}
Catastrophic forgetting poses a significant challenge to CTTA, and it primarily stems from inter-domain interference.
In this paper, we introduce domain-specific prompts that guide model adaptation, facilitating the partial disentanglement of parameter spaces across different domains.
As the domain identity for target samples is unknown, we use a query mechanism to dynamically infer if data comes from a previously seen or new domain, followed by prompt tuning via mutual information maximization and structural regularization.
Experiments on three benchmark datasets demonstrate the effectiveness of our PAINT method.
One limitation of our current study lies in the assumption that the source and target domains share an identical category space, which may not hold in practice. 
Future work will explore extending PAINT to the open-set TTA setting~\cite{gao2024unified}, where target domains may contain samples from unknown categories.



\bibliographystyle{elsarticle-num}
\bibliography{PAINT_ref}


\end{document}